\documentclass{article}

  \usepackage[nonatbib,preprint]{neurips_2021}

\usepackage[sorting = none, backend = bibtex, style=numeric-comp]{biblatex}
\usepackage[utf8]{inputenc} % allow utf-8 input
\usepackage[T1]{fontenc}  % use 8-bit T1 fonts
\usepackage{hyperref}    % hyperlinks
\usepackage{url}      % simple URL typesetting
\usepackage{booktabs}    % professional-quality tables
\usepackage{amsfonts}    % blackboard math symbols
\usepackage{nicefrac}    % compact symbols for 1/2, etc.
\usepackage{microtype}   % microtypography
\usepackage{xcolor}     % colors
\usepackage{amsmath}

\usepackage{graphicx}
\usepackage[shortlabels]{enumitem}
\usepackage[normalem]{ulem}
\usepackage{soul}

\title{Unsupervised Reservoir Computing for Solving Ordinary Differential Equations}
\author{%
 Marios Mattheakis\thanks{https://scholar.harvard.edu/marios$\_$matthaiakis/home}, Hayden Joy, Pavlos Protopapas \\
John A. Paulson School of Engineering and Applied Sciences, Harvard University \\
Cambridge, Massachusetts 02138, United States \\
\texttt{mariosmat@seas.harvard.edu}, 
\texttt{hnjoy@mac.com}, 
\texttt{pavlos@seas.harvard.edu}\\
}

\newcommand{\uu}{ {\bf u} }
\newcommand{\hh}{ {\bf h} }
\newcommand{\HH}{ {\tilde {\bf h}} }
\newcommand{\HHH}{ {\bf H} }
\newcommand{\G}{ {\bf G} }

\newcommand{\Sb}{ {\bf S} }

\newcommand{\bb}{ {\bf b} }

\newcommand{\wout}{ {\bf W}_\text{out} }
\newcommand{\win}{ {\bf W}_\text{in} }
\newcommand{\wres}{ {\bf W}_\text{res} }
\newcommand{\rctorch}{ \texttt{rcTorch} }

\newcommand{\yy}{y}

\begin{document}

\maketitle

\begin{abstract}
There is a wave of interest in using unsupervised neural networks for solving differential equations.
The existing methods are based on feed-forward networks, {while} recurrent neural network differential equation solvers have not yet been reported. We introduce an unsupervised reservoir computing (RC), an echo-state recurrent neural network capable of discovering approximate solutions that satisfy ordinary differential equations (ODEs). 
We suggest an approach to calculate time derivatives  of recurrent neural network outputs  without using backpropagation. 
The internal weights of an RC are fixed, while only a linear output layer is trained, yielding efficient training. However, RC performance strongly depends on finding the optimal hyper-parameters, which is a computationally expensive process. We use Bayesian optimization to efficiently discover optimal sets in a high-dimensional hyper-parameter space and numerically show that one set  is robust and can be used to solve an ODE  for different initial conditions and time ranges.
A closed-form formula for the optimal output weights is derived to solve first order linear equations in a backpropagation-free learning process. 
We extend the RC approach by solving nonlinear system of ODEs using a hybrid optimization method consisting of gradient descent and Bayesian optimization. 
Evaluation of linear and nonlinear systems of equations demonstrates the efficiency of the RC  ODE solver.

%keys: non-autonomous systems, backpropagation-free learning, unsupervised recurrent network, differential equations
\end{abstract}

\section{Introduction}
Neural networks (NNs) have been widely applied recently to study various kinds of differential equations. Physics-informed NNs  can be trained on data to learn nonlinear differential operators \cite{deeONET}, discover differential equations  \cite{rudy2017scienceAdv,kutz2017data}, and find approximate solutions for those equations \cite{sinai2018}. These data-driven supervised networks have been applied to a variety of real-world problems such as
%learning induced biases like Hamiltonians  to generate realistic trajectories
learning the dynamics of mechanical systems  \cite{HNN_nips2019, chaos2019, HNN_PRE2020} and designing meta-materials for nano-photonics \cite{inverseOpticsOE2020}. %Physics-informed neural networks are able to discover induced-biases like Hamiltonians and Lagrangians, that leads to differential equations system able to describe ground truth data [Hamiltonain, Lagrangian, check the refs in pHNN and VIGN]. 
% \sout{Physics-informed networks have found great applications in robotic motion and control systems.} 
 Unsupervised  NNs  have been used to solve a variety of differential equations  such as ordinary differential equations (ODEs) \cite{lagaris1998, lydia2020, cedric2020, mariosHNN}, partial differential equations \cite{pnas2018, spiliopoulos2018, neuroDiff2020, mariosHNN}, and eigenvalue problems \cite{ Jin2020}; these networks do not use any labeled data. Semi-supervised models have been applied to learn general solutions of differential equations  and extract  solutions which best fit given data \cite{alessandro2020}. Unsupervised NN solvers present exceptional advantages over traditional integrators: they suffer less from the "curse of dimensionality" in solving high-dimensional partial differential equations \cite{pnas2018, grohs2018}, the numerical solutions are obtained in a closed and differentiable form \cite{lagaris1998}, numerical errors are not accumulated in the  solutions \cite{mariosHNN}, initial and boundary conditions are identically satisfied \cite{lagaris1998, Jin2020}, and the solutions can be inverted \cite{alessandro2020,inverseOpticsOE2020, raissi2019}. Despite the success of the aforementioned models, all approaches are based on feed-forward NNs, while, to the best of our knowledge, recurrent neural networks (RNNs) for solving differential equations in an unsupervised fashion have not been reported yet. In this study, we fill the gap by introducing an unsupervised RNN, in the context of reservoir computing (RC) \cite{Jaeger2004}, which is able to discover approximate solutions to systems of ODEs. 

RC is an echo state RNN where the internal parameters (weights and biases) are fixed, while only a linear output layer needs to be trained yielding fast and computationally efficient training \cite{Jaeger2004, Ott2017, mariosFrontier2019}. Fixing the internal weights eliminates gradient exploding/vanishing problems during the training \cite{Jaeger2004}. 
RC has been widely used for studying dynamical systems such as weather forecasting \cite{prl2018_ott}, predicting chaotic \cite{Ott2017, RC_chaoticSeries_nips2020} and irregular behavior \cite{mariosFrontier2019}, and classifying time series \cite{RC_classification_nips2020}.
Moreover, the RC architecture has been adopted to build physical hardware NNs for neuromorphic computing \cite{RC_neuromorphic_nips2012,scRep8_2018, Tanaka2019RecentAI_neuromorphicRC, scRep9_2019, prx7_2020, prl_2020}. 
%In a physical RC network, the hidden states are harvested by natural processes like electromagnetic wave dynamics in a photonic device, that is, the reservoir is implemented in digital hardware. Then, a readout output layer is trained to perform a certain task, where the output weights are usually implemented in software. 
Neuromorphic devices are currently being developed for intelligent and energy-efficient devices providing extremely fast real-time computing with very low energy cost. This study suggests an avenue to design neuromorphic differential equation solvers.

The proposed RC solver is an extension to NN differential equation solvers and consequently, acquires all the benefits that network solvers have over numerical integrators.
 Moreover, RNNs perform better than feed-forward NNs on sequential data. Considering that the solutions of ODEs are time series and, therefore, similar to sequential data, we expect RNNs to generalize better than feed-forward NNs. 
 %However, training RNNs is a very slow process and usually suffers from exploding and vanishing gradient problems. To avoid the aforementioned issues, in this study we use a special class of RNN called RC, where
 The internal weights of an RC are fixed and randomly initialized from distributions that have certain statistical properties which yield the echo-state property \cite{Jaeger2004, Ott2017}. 
Training an RC is  efficient, however, finding the optimal hyper-parameters that determine the weight distributions and the network architecture is challenging.
%since the hyper-parameters lie in a high-dimensional space.
RC performance strongly depends on these hyper-parameters and, therefore, finding good hyper-parameters is crucial. The RC has many hyper-parameters making this a computationally expensive process. 
Conventionally, simple grid-search in the hyper-parameter space is used \cite{Ott2017}, however, this method is prohibitively expensive for more than three hyper-parameters. 
Tuning hyper-parameters by using Bayesian Optimization (BO) is an active field of research \cite{unk2019BO, TURBO2020, Maat2018, dai2020federated, Botorch2020}. In this study, we use TURBO-1, a BO method introduced in \cite{TURBO2020}. The strength of this method is discussed in the supplementary material (SM). We find that a single set of hyper-parameters is robust and can be used to solve  ODEs for different initial conditions and time ranges. 

% {\bf Our contributions is fourfold}:
{\bf Main contributions}: First, in the continuous time limit we suggest an approximation to calculate  time derivatives of the outputs of an RNN without using backpropagation, which is a substantial ingredient in ODE-solver neural networks. Second, we introduce an unsupervised RC and show that it is capable of solving ODEs. Third, we derive a closed-form formula for the RC trainable parameters for solving first order non-autonomous linear ODEs in a backpropagation-free learning method. Fourth, we develop a hybrid method consisting of gradient descent and BO which is able to find optimal hyper-parameters and weights of an RC for solving systems of nonlinear ODEs. For this study, we use the \rctorch library, an RC framework written in pytorch with embedded BO that utilizes BoTorch \cite{Botorch2020}. A public open-source github repository is available at \hyperlink{https://github.com/blindedjoy/RcTorch}{https://github.com/blindedjoy/RcTorch}.

%\mm{Add the outline/structure of the paper}

% ######################################################
% \section{Background: Continuous-time recurrent neural networks and hyper-parameters optimization}

\section{Background: Neural network differential equations solvers}
%As it is mentioned in the introduction, unsupervised neural networks have been widely used for solving differential equations.
% Neural networks have been widely used for solving  ordinary \cite{mariosHNN, lagaris1998}, partial \cite{spiliopoulos2018, neuroDiff2020}, and eigenvalue \cite{Jin2020} differential equation problems while they suffer less than traditional numerical methods from the "curse of dimensionality" \cite{pnas2018, grohs2018}.
%
Several software libraries of NN differential equations solvers have been recently developed, including NeuroDiffEq \cite{neuroDiff2020}, DeepXDE \cite{deepXDE}, and SimNet \cite{simnet2020}, indicating that developing NN solvers is an active area of research \cite{karniadakisNatureReview2021}.
All of these libraries are based on feed-forward NNs and developed with the pytorch or tensorflow software platforms, where the automatic differentiation mechanism is employed to compute analytical derivatives used to define the loss function.
In this context NN  solvers are unsupervised learning methods. Since we want to solve differential equations, we do not know the corresponding target solutions and thus, we lack labeled or ground truth data. The only accessible information is the differential equation and the associated initial or boundary conditions. The loss function solely depends on the network predictions including the differential equation, while the initial/boundary conditions are embedded in the structure of the network and are thus identically satisfied. 
Neural network solvers are able to solve ordinary and partial differential equations of an arbitrary order. We nevertheless focus on ODEs in this study — particularly on systems of first order ODEs since higher order ODEs can be decomposed into systems of first order ODEs. We review here the approach of developing NNs for solving a first order ODEs subjected to certain initial conditions (ICs). Consider a general  ODE of the form
\begin{align}
  \label{eq:diffOperator}
D\left(t,\psi, \dot \psi \right) -f(t) = 0,
\end{align}
where $t$ is the independent variable time, and $\psi(t)$ is the dependent variable subjected to a certain IC $\psi(0)=\psi_0$. In Eq. (\ref{eq:diffOperator}), $D$ is an arbitrary function of $\psi(t)$ and its first time derivative $\dot \psi$, and $f(t)$ is a forcing function of $t$. Equation (\ref{eq:diffOperator}) describes a non-autonomous system since it explicitly dependents on time. Considering known $ D$ and $f$ functions, we are seeking $\psi(t)$ that solves Eq. (\ref{eq:diffOperator}) and  satisfies a given $\psi_0$. Specifically, the goal is to construct a numerical solution $\yy$ which approximates the unknown ground truth solution $\psi$. To achieve that, we employ a NN  that takes an input $t=(t_1, t_2, \dotsc, t_K)$ and returns an output $N(t,p)$; $K$ indicates the total number of the input data points, and $p$ denotes the trainable parameters of the network. 
An efficient way to impose ICs was introduced in Ref. \cite{lagaris1998} and suggests using of a parametric solution $\yy(t)$ of the form:
\begin{align}
  \label{eq:parSol0}
  \yy(t) &=\psi_0 + g(t) N(t,p) 
\end{align}
where $g(t)$ can be any arbitrary function of $t$ with the constraint $g(0) = 0$. The parametrization of Eq. (\ref{eq:parSol0}) generates NN solutions that identically satisfy the ICs \cite{lagaris1998, mariosHNN, lydia2020}, namely $\yy(0)=\psi_0$. 
Having the parametric solution of Eq. (\ref{eq:parSol0}), which is a function of $N$, solving the ODE of Eq. (\ref{eq:diffOperator}) is reduced to an optimization problem of the form
\begin{align}
  \label{eq:loss0}
  \underset{p}{\arg\min} \left( \sum_{n=0}^{K}
  \Big( D\left(t_n, \yy_n, \dot \yy_n \right) -f(t_n) \Big)^2 \right), 
\end{align}
where $\yy_n = y(t_n)$. %, $\dot \yy_n = \dot y(t_n)$, and $\ddot \yy_n = \ddot y(t_n)$. 
The sum in Eq. (\ref{eq:loss0}) defines a loss function whose minimization yields  $p$ that constructs a neural solution $y$ which approximately solves the  ODE  (\ref{eq:diffOperator}). % and identically satisfies the given ICs $\psi_0$.
It is worth noting that $\yy(t)$ can approximate $\psi(t)$ with arbitrary small error due to the universal approximation theorem of NNs  \cite{hornik1991}. 
Furthermore, regularization terms can be used  to penalize large weights  or to impose physical principles like energy conservation \cite{mariosHNN}.
%, or to encourage the network to search for different sets of eigenvalues and eigenfunctions \cite{Jin2020}.
Subsequently, the total loss function is expressed as:
\begin{align}
  L &= L_\text{ODE} + L_\text{reg} \nonumber \\
   \label{eq:loss1}
   &= \sum_{n=0}^{K}
   \Big(D\left(t_n, \yy_n, \dot \yy_n \right) -f(t_n) \Big)^2  + L_\text{reg}.
\end{align}
 
The  NN  solution $y$ of Eq. (\ref{eq:parSol0}) is a closed-form solution, meaning that it can be evaluated at every time point, differentiated, and inverted. These are unique properties of NN  solvers that are not shared by standard integrators. Despite the advantages shown by feed-forward NN solvers, an RNN solver is missing from the literature. In this work, we present a novel RNN solver, specifically an echo-state RC, capable of solving ODEs in a given training and IC range.

\section{Continuous-time recurrent neural networks}

Learning from sequential data is a challenging task for machine learning because of the underlying time correlation. RNNs share parameters across the hidden layers giving them an intrinsic memory and subsequently, they are well suited to handle sequential data.
Standard RNNs require discrete input data at discrete time points. In the continuous-time limit, when the discrete time points are close to each other, the dynamics of the hidden RNN layers can be approximated by continuously defined dynamics through ODEs \cite{resNet_Ode_AAAI2018, neuralODE_nips2018, hyperSolvers_nips2020}. This approximation has been adopted by residual networks \cite{resNet_Ode_AAAI2018} and continuous depth models \cite{neuralODE_nips2018, hyperSolvers_nips2020}. This is a core idea in the present study because it allows, in a backpropagation-free process, the calculation of the time derivatives of the outputs of an RNN.

%\subsection{Calculating derivatives in recurrent neural networks}

Consider an RNN unit with $P$ input time series   $\uu(t)=(u_1(t), \dotsc u_P(t)) \in {\rm I\!R}^P$, and  $M$ hidden recurrent neurons described by a temporal state vector $\hh(t) = (h_1(t), \dotsc h_M(t)) \in {\rm I\!R}^{1\times M}$, where  the time variable consists of  $K$  points as $t=(t_1, t_2, \dotsc, t_K)$.
We show that in the continuous-time limit where the time step $\Delta t$ between two sequential data points is very small ($\Delta t \ll 1$), the dynamics of a leaky RNN unit can be approximated by a system of first order nonlinear ODEs  of the form,
\begin{align}
\label{eq:hDE}
\dot \hh = -\tilde \alpha \hh + \tilde \alpha f\left( \wres \cdot \hh +\win\cdot \uu +\bb \right),
\end{align}
where $\dot \hh \equiv d \hh/dt$ and the dot denotes the inner product. The input weights and biases are represented, respectively, by $\win\in {\rm I\!R}^{M \times P}$ and $\bb\in {\rm I\!R}^{M}$, $\wres\in {\rm I\!R}^{M\times M}$ describes the recurrent weights, $\tilde \alpha$ is the leakage rate, and $f(\cdot)$ denotes a nonlinear activation function \cite{Ott2017}. 
Applying a Euler discretization for the first derivatives, $\dot \hh = \frac{\hh_{n+1} - \hh_n}{\Delta t}$, the system of Eq. (\ref{eq:hDE}) takes the discrete form 
\begin{align}
\label{eq:hn}
\hh_{n+1} = \left(1-\alpha \right)\hh_{n} + \alpha f\left( \wres \cdot \hh_{n} +\win\cdot \uu_n +\bb \right),
\end{align}
with $\alpha = \tilde \alpha \Delta t$. Equation (\ref{eq:hn}) describes the update of a leaky RNN unit and subsequently, it is a first order approximation of the continuous model described by Eq. (\ref{eq:hDE}). Since Eq. (\ref{eq:hn}) determines all $\hh_n$ of an RNN, Eq. (\ref{eq:hDE}) provides, without any computational cost, the first time derivatives as:
\begin{align}
\label{eq:hn_dot}
\dot \hh_n = -\frac{\alpha}{\Delta t} \hh_n + \frac{\alpha}{\Delta t} f\left( \wres \cdot \hh_n +\win\cdot \uu_n +\bb \right).
\end{align}
%We note that the derivative in Eq. (\ref {eq:hn_dot}) is an approximation in the context of the firs
Higher order derivatives can be calculated by taking time derivatives of Eq. (\ref{eq:hn_dot}) and applying the chain rule to the $f(\cdot)$ term.
%Accordingly, taking a time derivative of Eq. (\ref{eq:hn_dot}), we calculate the second time derivative of the hidden states, 
%\begin{align}
%\label{eq:hn_ddot}
%\ddot \hh_n = -\frac{\alpha}{\Delta t} \dot \hh_n + \frac{\alpha}{\Delta t} \frac{\partial f}{\partial t} \left( \wres \cdot \dot \hh_n +\win\cdot \dot \uu_n  \right),
%\end{align}
%where $\dot \hh$ is given be Eq. (\ref{eq:hn_dot}). Following the some procedure, we \hj{\st{have access to} can obtain the} higher order derivatives. \hj{\st{We note that}} 
%Calculating the second and higher time derivatives of the hidden states requires time derivatives of the input sequences $\uu$ which may not be available. On the other hand, the first derivative is always available since Eq. (\ref{eq:hn_dot}) does not contain any time derivatives of $\uu$. Moreover, it is worth noting that there is no computational cost in calculating the derivatives since we do not need to use numerical differentiation or back-propagation. 
The only numerical error in the first derivative of Eq. (\ref{eq:hn_dot}) is introduced through the assumption that Eq. (\ref{eq:hn}) is derived from Eq. (\ref{eq:hDE}) by applying a Euler discretization. This error can be arbitrary small by appropriately choosing a small $\Delta t$. 
%Further numerical error is not introduced in the higher derivatives. %, higher derivatives like Eq. (\ref{eq:hn_ddot}) can be calculated exactly by computing the analytical derivative. 
Consequently, the time derivatives of the hidden states can be estimated  without using backpropagation. 

Considering the general case of an RNN that returns $R$ outputs $N(t)\in {\rm I\!R}^{R} $, we read
\begin{align}
  \label{eq:N_rnn}
  N(t_n) = {\bf W_o} \cdot \hh_n +{\bf b_\text{o}},
\end{align}
where ${\bf W_o} \in {\rm I\!R}^{R\times M}$ and ${\bf b_\text{o}} \in {\rm I\!R}^{R}$ are the weights matrix and biases of an output linear layer. The time derivative of $N$ can be calculated in a backpropagation-free mode using the result in Eq. (\ref{eq:hn_dot}) as:
\begin{align}
  \label{eq:Ndot_rnn}
  \dot N(t_n) = {\bf W_o} \cdot \dot \hh_n.
\end{align}
Backpropagating in RNNs is  computationally expensive and can be  impractical  for large $K$. On the other hand, through  Eq. (\ref{eq:Ndot_rnn})   we can compute time derivatives without computational cost. Although throughout this study we apply Eqs. (\ref{eq:hn_dot}) and (\ref{eq:Ndot_rnn}) for the RC architecture, the approach has broader implications since it holds for any RNN. Subsequently, it opens the door to a wide range of potential applications including general differential equation NN solvers and physics-informed RNNs.

%Accordingly, we will see later that have the derivative of the hidden states yields to the time derivatives of the output of an RNN. This is a very useful result because in RNN the backpropagation is computationally expensive and for large $K$ can be practically prohibited. Although in this study we apply Eq. (\ref{eq:hn_dot}) for the RC architecture, the approach has a broader impact since it holds for any RNN opening the door to a wide range of applications including general differential equation solvers and physics-informed RNNs.

\section{Reservoir computing forms an ordinary differential equation solver}

%Supervised RC has been used on a wide variety of data-driven scientific problems such as forecasting the behavior of chaotic systems \cite{RC_chaoticSeries_nips2020} and time-series classification \cite{RC_classification_nips2020}. To our knowledge no applications of unsupervised RC have been reported to date. 
In this section, we introduce an unsupervised RC model that takes $t$ as an input sequence, namely $\uu_n = t_n$, and is trained to solve ODEs within the range of $t$. First, we examine single linear and nonlinear first order ODEs. Later, we modify the proposed RC to solve systems of first order ODEs. 
Similar to feed-forward NN  solvers, the objective of the proposed machine learning method is to minimize the loss function of Eq. (\ref{eq:loss1}) for   given  ODE  and  ICs.

We employ an RC that  returns one output sequence $N$, hence Eqs. (\ref{eq:N_rnn}) and  (\ref{eq:Ndot_rnn}) yield
\begin{align}
  \label{eq:N}
  N(t_n,\wout) &= \wout \cdot \HH(t_n), \\
    \label{eq:Ndot}
  \dot N(t_n,\wout) &= \wout \cdot {\dot{\HH}}(t_n), 
\end{align}
where $\HH=[1, \hh]$ contains a column of ones accounting for the bias of the output layer of the RC,  $\dot \HH = [0, \dot \hh]$ since the constant bias vanishes after operating the derivative, and the readout (output) layer  $\wout \in {\rm I\!R}^{R\times (M+1)}$ accounts for the only trainable (weights and bias) parameters of the RC.
Using the parameterization in Eq. (\ref{eq:parSol0}), we construct the  RC solution 
\begin{align}
  %\yy(t,\wout) &=\psi_0 + g(t) N(t, \wout) \nonumber \\
   \label{eq:parSol}
   \yy(t,\wout) &=\psi_0 + g(t) \wout \cdot \HH(t),
\end{align}
with $g(0) = 0$ and thus, $\yy(0,\wout)=\psi_0$.
Having the RC parametric solution of Eq. (\ref{eq:parSol}) we  train  $\wout$ such that Eq. (\ref{eq:loss0}) is minimized and subsequently, we obtain $\yy$ that approximately satisfies the general ODE  (\ref{eq:diffOperator}) and identically satisfies the given ICs.

The $L$ of Eq. (\ref{eq:loss1}) can be  minimized using gradient descent and backpropagation through a linear layer. 
%Since all trainable parameters of an RC appear at the output linear layer, $L$ is a convex function and thus,  training $\wout$ with gradient descent is efficient.  
Interestingly, for linear non-autonomous first order ODEs, a closed form solution of $\wout$ that minimize Eq. (\ref{eq:loss1}) is  derived and thus, numerical optimization is not required. Consequently,  solving linear ODEs  with RC is a backpropagation-free training method.

% \section{Single non-autonomous ordinary differential equations}
% In this section we employ RC to solve single non-autonomous first order ODEs. 

%\subsection{Linear differential equations: Backpropagation-free learning method }
{\bf Linear differential equations, backpropagation-free learning method: }
In the RC architecture, the only adjustable parameters appear in the output layer, giving the opportunity to get an analytical closed-form solution for the optimal $\wout$. We exploit this potential by studying first order linear non-homogeneous ODEs. These equations often appear in diffusion processes like fluid dynamics, and are  described by the general linear differential equation: %operator 
\begin{align}
%   D^{(1)} (t,\psi,\dot \psi) -f(t) &=0 \nonumber \\
  \label{eq:ODE1}
   a_1(t) \dot \psi + a_0(t) \psi - f(t)&=0,
\end{align}
where the coefficients $a_0(t)$, $a_1(t)$ and the force $f(t)$ are continuous functions of $t$.
%
%We calculate approximate solutions $\yy$ of the general ODE (\ref{eq:ODE1}). 
Minimizing Eq. (\ref{eq:loss1}) for the ODE Eq. (\ref{eq:ODE1}) when $\yy$ is used instead of $\psi$,  a closed-form solution for $\wout$ is derived to produce  $\yy$ that approximately solves the ODE (\ref{eq:ODE1}).
 Substituting   Eq. (\ref{eq:parSol}),  Eq. (\ref{eq:ODE1}) can be elegantly re-expressed in matrix notation as
\begin{align}
\label{eq:ODE1_matrix}
A_1 \dot Y + A_0 Y -F = 0,
\end{align}
with $F = \big(f(t_0), \dotsc, f(t_{K})\big)^T$, $A_i =\big(a_{i}(t_0), \dotsc, a_i(t_{K})\big)^T$ $(i=0,1)$, and the RC solution  $Y = \big(y(t_0,\wout), \dotsc, y(t_{K},\wout) \big)^T$ % that is compactly written according to Eq. (\ref{eq:parSol}) as: 
%In the matrix representation the parametric network solution in
which is written according to Eq. (\ref{eq:parSol}) as:
\begin{align}
  Y &= \Psi_0 + \left( \G \circ \HHH \right) \wout \nonumber \\
  \label{eq:parSol1}
  &= \Psi_0 + \Sb \wout,
\end{align}
and the associate time derivative reads:
\begin{align}
  \label{eq:parSol1_dot}
  \dot Y= \dot \Sb \wout,
\end{align}
where $\circ$ denotes the Hadamard product, $\Psi_0 \in {\rm I\!R}^{ K}$ is the constant vector $\Psi_0=(\psi_0, \dotsc, \psi_0)^T$, $\G\in {\rm I\!R}^{K\times M}$ is a matrix with repeating rows of   $\left(g(t_0), \dotsc, g(t_{K})\right)^T$,  $\HHH\in {\rm I\!R}^{K\times M}$ is the  state matrix $\HHH=\left( \HH(t_0), \dotsc, \HH(t_{K}), \right)^T$, and $\Sb=\G \circ \HHH \in {\rm I\!R}^{K\times M}$. To derive a close-form solution of $\wout$ we consider  $L_2$ regularization, $L_\text{reg} = \lambda \wout^T \wout$, where $\lambda$ is the regularization parameter \cite{Ott2017}.
% \begin{align}
%   \label{eq:ridge}
% L_\text{reg} = \lambda \wout^T \wout,
% \end{align}
% where $\lambda$ is the ridge parameter.
Minimizing  $L$ of Eq. (\ref{eq:loss1}) for the ODE of Eq. (\ref{eq:ODE1_matrix}) and with $L_2$ regularization, we get:
%and $T$ denotes the transpose of a matrix.
\begin{align}
  \frac{\partial }{\partial \wout}\Big[ \left(A_1 \dot Y + A_0 Y -F\right)^T \left(A_1 \dot Y + A_0 Y -F\right) + \lambda \wout^T \wout \Big] &=0 \nonumber \\
  \left(A_1 \dot \Sb \wout + A_0\Sb\wout + A_0\Psi_0 -F \right)^T\left(A_1\dot \Sb + A_0\Sb\right)+\lambda\wout^T &=0
  \nonumber \\
  \label{eq:minLoss1}
  \left( \wout^T D_\HHH^T +D_0^T\right)D_\HHH +\lambda\wout^T &=0,
  %\wout^T\left( A_1 \dot \Sb^T\Sb + A_0 \Sb^T\Sb+\lambda\mathbf{1}\right) &= -\left(A_0\right)
\end{align}
where we define the matrices
\begin{align}
  \label{eq:DH}
  D_\HHH &= A_1\dot \Sb + A_0 \Sb, \\
  \label{eq:D0}
  D_0 &= A_0 \Psi_0 - F.
\end{align}
Solving Eq. (\ref{eq:minLoss1}) for $\wout$  yields a closed-form equation of $\wout$ that constructs an RC solution of Eq. (\ref{eq:parSol1}) which approximately solves any linear non-homogeneous first order ODE, hence
 \begin{align}
   \label{eq:analyticW}
   %\wout &= -\left(D_\HHH^T D_\HHH\right)^{-1} D_\HHH^T D_A.
   \wout = - \left(D_\HHH^T D_\HHH + \lambda{\bf 1} \right)^{-1}D_\HHH^T D_0,
 \end{align}
where ${\bf 1}$ is a $1 \times M$ vector of ones.
 We read that Eq. (\ref{eq:analyticW}) consists of two characteristic matrices, $D_\HHH$ and $D_0$ given by Eqs. (\ref{eq:DH}) and (\ref{eq:D0}), respectively. The former ($D_\HHH$) contains information of the RC hidden states $\HHH$, while the last  ($D_0$) includes the ICs and the force function. Both characteristic matrices are informed about the differential equation since the coefficients $A_0,~A_1$ appear in both places.
We are able to obtain the closed-form solution $\wout$ because $Y$ and $\dot Y$ of Eqs. (\ref{eq:parSol1}), (\ref{eq:parSol1_dot}) are both linear to $\wout$. Thus, their first derivative with respect to $\wout$ is independent of $\wout$, and therefore, a linear system for $\wout$ is derived in Eq. (\ref{eq:minLoss1}). Such a closed-form is not possible for nonlinear ODEs since a linear system of $\wout$ cannot be derived with the parametrization of Eq. (\ref{eq:parSol1}).
 
 Equation (\ref{eq:analyticW}) states that the RC solver can be trained to solve linear first order ODEs without using numerical optimization, such as gradient descent and backpropagation. 
 The computationally costly part of the proposed RC solver is the hyperparameter optimization and this is the reason that an efficient method such as BO \cite{Maat2018} is crucial. Through  numerical experiments, we demonstrate that one set of hyper-parameters is sufficient for a wide range of ICs. Using the same set means that all the RC solutions for a particular ODE share the same states. Subsequently, we construct \HHH  for  one IC and  reuse them  for  additional ICs  allowing a computationally efficient  exploration of many  ICs.

%%%%%%%%%%%%%%%%1ST ORDER ODE%%%%%%%%%%%%%%%%i
% \subsubsection{Experimental evaluation}
%\subsection{First order linear ODEs}%: Theory and experimental evaluation}

\begin{figure}[ht!]
  \centering
\includegraphics[scale=0.33]{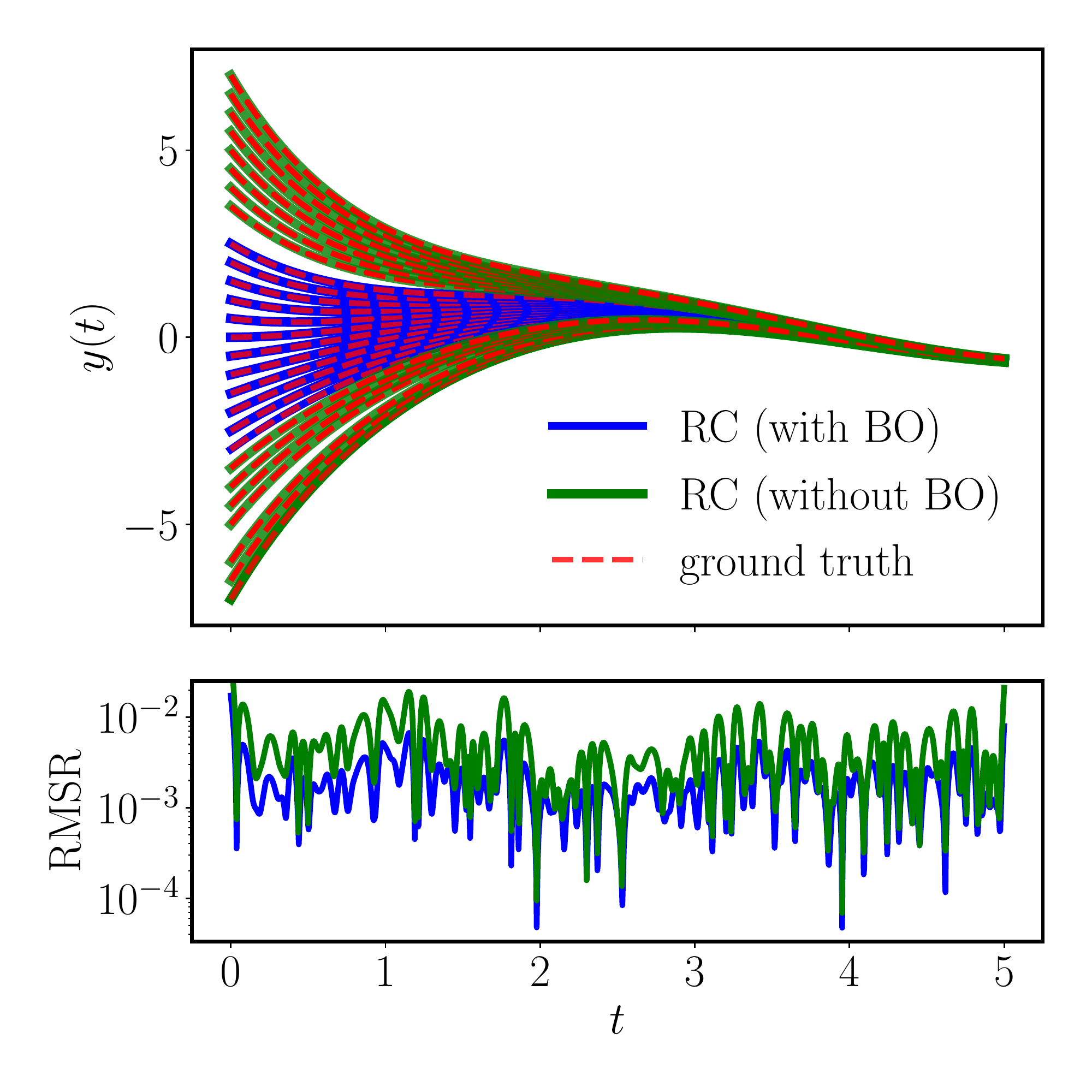}
  \includegraphics[scale=0.33]{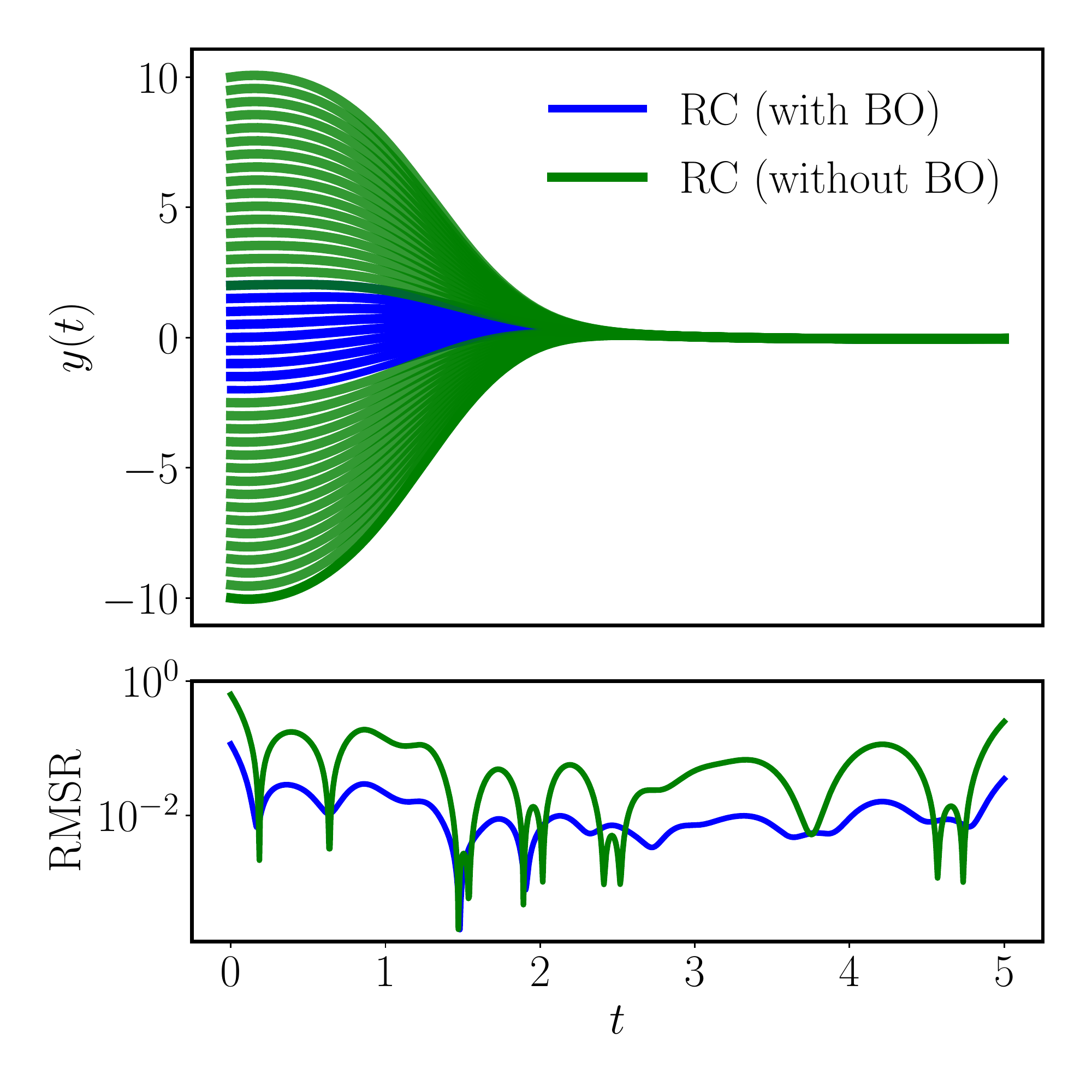}
  \caption{Backpropagation-free training for solving the  linear ODEs of Eq. (\ref{eq:ODE1_exp}) (right) and Eq. (\ref{eq:ODE1_b}) (left). Upper graphs outline the RC solutions for different ICs. Lower panels show the RMSR.  Blue and green lines represent ICs used and not used in BO. }% BO for a single IC. In the residuals plot we can show the residuals for BO on a single and on a bundle of ICs.}
  \label{fig:ode1_linear}
\end{figure}

%{\bf Experimental evaluation:}
We evaluate the closed-form solution of Eq. (\ref{eq:analyticW}) by solving two linear ODEs for different ICs. During experimental evaluation we adopt the efficient parametric function used in Refs. \cite{mariosHNN, lydia2020}:
\begin{align}
  \label{eq:parFun_exp}
  g(t) = 1 - e^{-t}.
\end{align}
We assess the RC performance by calculating the root-mean-square-residuals (RMSR) of the RC solutions, namely
\begin{equation}
  \label{eq:RMSR}
  \text{RMSR}(t) = \sqrt{ \frac{1}{L} \sum_\text{ICs} \Big( a_1(t) \dot{\yy} + a_0(t) \yy -f(t) \Big)^2},
\end{equation}
where the sum denotes averaging along $L$  different $\psi_0$.
For the first experiment we consider the ODE
\begin{align}
  \label{eq:ODE1_exp}
  \dot \psi + \psi = \sin(t),
\end{align}
which has the exact solution:
\begin{align}
\label{eq:exactSolutionODE1a}
\psi(t) = e^{-t}\left(y_0 + \frac{1}{2}\right) + \frac{1}{2}\left( \sin(t) - \cos(t) \right).
\end{align}
From Eq. (\ref{eq:ODE1_exp}) we note that $a_0=a_1=1$ and $f(t)=\sin(t)$. We get the optimal $\wout$ by calculating the characteristic matrices of Eqs. (\ref{eq:D0}), (\ref{eq:DH}) and substituting them into Eq. (\ref{eq:analyticW}). Then, we construct the RC solution by employing Eq. (\ref{eq:parSol}). % and the parametric function in Eq. (\ref{eq:parFun_exp}). 
The RC solutions of Eq. (\ref{eq:ODE1_exp}) along with the exact solutions (\ref{eq:exactSolutionODE1a}) are demonstrated in the left side of Fig.  \ref{fig:ode1_linear} for several ICs in the range $\psi_0=[-5.5,5.5]$. Upper graph: the solid blue lines account for the RC solutions while the dashed red curves indicate the exact solutions; each pair of solid-dashed lines corresponds to a solution with different $\psi_0$. The lower image shows the RMSR. The blue color indicates the ICs used in the BO to obtain the optimal hyper-parameters, while for  the solutions indicated by green lines we apply only the exact $\wout$. 
%These results are evidence that a set of hyperparameters can be used to construct solutions for different ICs of the same ODE. 

The second numerical experiment is an ODE with time dependent coefficients, defined by:
\begin{align}
  \label{eq:ODE1_b}
  \dot \psi + t^2 \psi = \sin(t),
\end{align}
where $a_0=t^2$, $a_1=1$, and $f=\sin(t)$. We calculate the optimal $\wout$ with Eq. (\ref{eq:analyticW}) and construct RC solutions  in the range of ICs $[-10,10]$. The RC predictions are shown in the right panel of Fig. \ref{fig:ode1_linear}. The upper graph demonstrates the predicted trajectories, while the lower image outlines the RMSR. There is no exact solution for the Eq. (\ref{eq:ODE1_b}), hence only the RC predictions are shown in the upper graph. We employ BO only for a few ICs shown in blue.

In both experiments, BO was applied to a bundle of ICs implying that a single set of hyper-parameters is sufficient to construct different RC solutions of the same ODE. In particular, applying BO and using the exact $\wout$ of Eq. (\ref{eq:analyticW}), we get the optimal hyperparameters that yielding the RC solutions shown with blue in Fig. \ref{fig:ode1_linear}. Then, using the same hyperparameters we exactly solve (without BO) for $\wout$  to construct RC solutions for different ICs as they are indicated in green in Fig. \ref{fig:ode1_linear}.
Moreover, the results shown in Fig. \ref{fig:ode1_linear} validate the closed-form solution of Eq. (\ref{eq:analyticW}) verifying that the RC solver is a backpropagation-free unsupervised machine learning method for solving linear first order ODEs.

% ######--------------------------

%\subsection{Nonlinear differential equations: Training with gradient descent}
{\bf Nonlinear differential equations, training with gradient descent: }
It is not possible to derive a closed-form solution for $\wout$ for nonlinear ODEs. Nevertheless, RC can solve nonlinear ODEs by training $\wout$ using gradient descent (GD). %, which is computationally efficient for the RC since we optimize a convex loss function.
We demonstrate the capacity of RC in solving nonlinear equations by studying Bernoulli type nonlinear equations of the form:
\begin{align}
  \label{eq:bernoulli}
  a_1(t) \dot y + a_0(t) y + q(t) y^2 = f(t).
\end{align}
Although it is not possible to derive an exact solution for the optimal $\wout$, an approximate closed formula is obtained through a linearization procedure. 
Then, we use the linearized $\wout$ instead of random weights to start the GD. This is a transfer learning approach that drastically accelerates and improves the training (see SM  for more details). 
In the context of linearization approximation  any nonlinear term of $\Sb \wout$ is dropped. This is a valid approximation since $\wout \ll 1$ due to the $L_2$ regression and $\Sb$ varies in the range $[-1,1]$ due to the parametric function of Eq. (\ref{eq:parFun_exp}) and the activation functions $\tanh()$ or $\sin()$, which are all bounded within $[-1,1]$. Consequently, we read $\wout \Sb \ll 1$ and thus, higher orders can be neglected, namely $(\wout \Sb)^\nu \simeq 0 $ for any integer $\nu>1$.

Minimizing the loss function of Eq. (\ref{eq:loss1}) for the nonlinear ODE (\ref{eq:bernoulli}) %and dropping nonlinear terms of $\wout$ (linearization),
yields
\begin{align}
  \left(A_1 \dot Y + A_0 Y +Q Y^T Y -F\right)^T  \frac{\partial }{\partial \wout} \left( A_1 \dot Y + A_0 Y +Q Y^T Y -F\right) + \lambda \wout^T  &=0 \nonumber \\
   \label{eq:minLoss_nl0}
   \Big( \wout^T D_\HHH^T +D_0^T + Q\left( \Psi_0^2 + 2 \Psi_0\wout^T \right) \Big)\Big( D_\HHH + 2Q\left(\Psi_0+\Sb\wout\right) \Big) +\lambda\wout^T &=0 \\
   \label{eq:minLoss_nl}
   \wout^T\Big( \tilde D_\HHH^T \tilde D_\HHH + \tilde 2 Q D_0^T \Sb + \lambda{\bf 1} \Big) + \tilde D_0^T \tilde D_\HHH &=0, 
\end{align}
where in Eq. (\ref{eq:minLoss_nl0}) nonlinear terms of $\Sb\wout$ are dropped.  $Q=(q(t_0),\dotsc, q(t_K) )$, and the modified characteristic matrices are defined as:
\begin{align}
  \label{eq:DH_nl}
  \tilde D_\HHH &= D_\HHH + 2 Q \Psi_0, \\
  \label{eq:D0_nl}
  \tilde D_0 &= D_0 + Q \Psi_0^2.
\end{align}
The linear algebraic system of Eq. (\ref{eq:minLoss_nl}) can be inverted to give the linearized RC weights as:
\begin{align}
\label{eq:wout_nl}
\wout = - \Big( \tilde D_\HHH^T \tilde D_\HHH + \tilde 2 Q D_0^T \Sb + \lambda{\bf 1} \Big)^{-1} \tilde D_\HHH^T \tilde D_0. 
\end{align}

\begin{figure}[h!]
  \centering
    \includegraphics[scale=0.4]{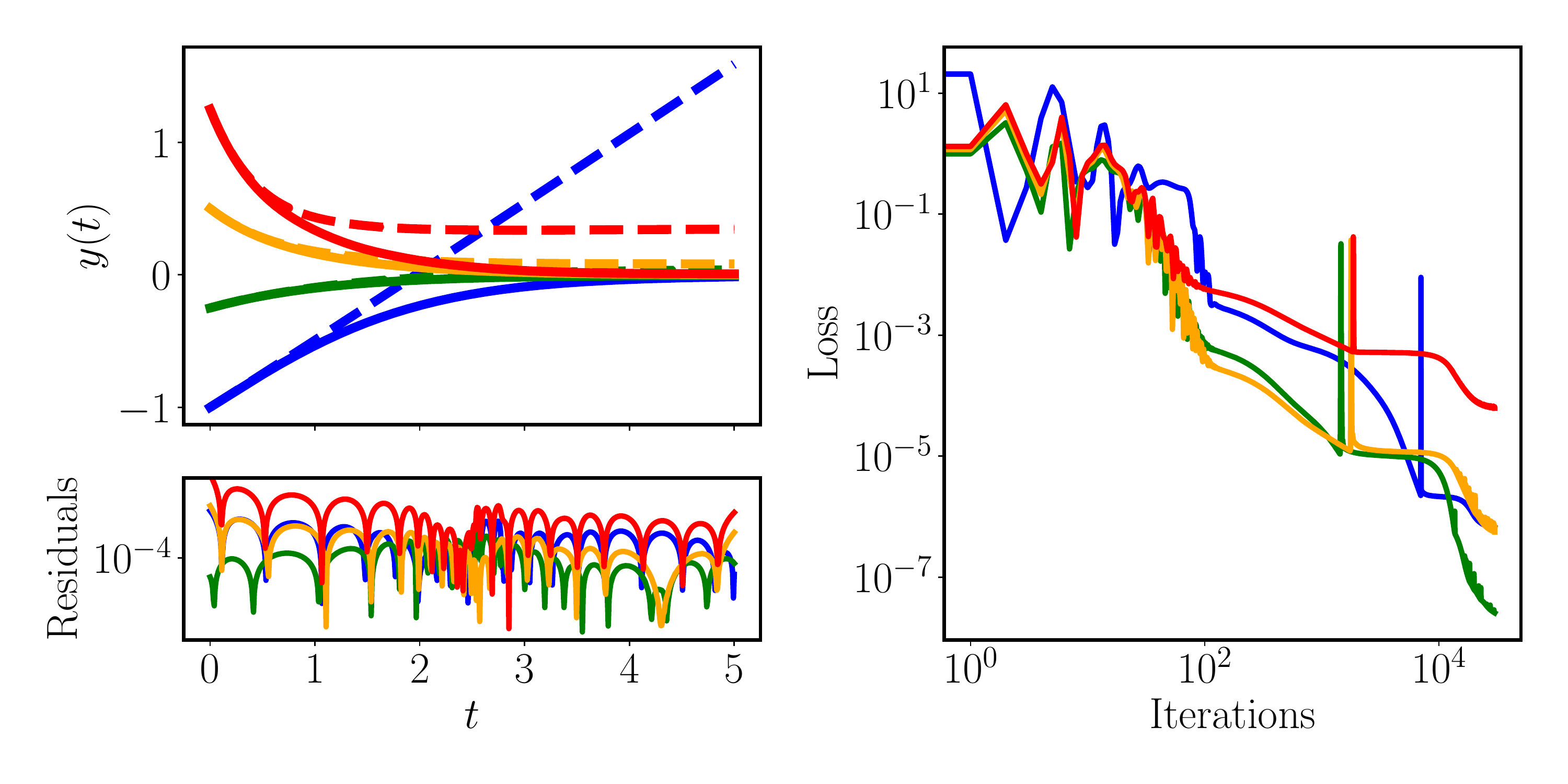}
  \caption{Nonlinear ODE. The upper-left image shows RC solutions obtained by linearized weights (dashed lines) and the solutions obtained after gradient descent optimization (solid lines). The lower-left panel outlines the residuals for each RC solution  after applying GD. The graph on the right shows the loss function during the GD iterations. Each color  represents a different IC.}
  \label{fig:bernoulli}
\end{figure}

 We assess the performance of the RC  by solving the  ODE  (\ref{eq:bernoulli}) for $a_0=a_1=1$, $f=0$, and $q=0.5$. Starting with the linearized $\wout$ of Eq. (\ref{eq:wout_nl}), we employ GD to train the parameters. This is efficient since we only  optimize a single linear layer.  Figure \ref{fig:bernoulli} presents the RC solutions (top-left graph) and the associated residuals (bottom-left) for different ICs indicated by different colors.  Upper plot: the dashed lines indicate  RC predictions obtained by solely applying the linearized $\wout$ of Eq. (\ref{eq:wout_nl}),  before applying GD; solid lines are the RC predictions after GD. 
 %We observe that for smaller ICs the linearized solutions are close to final solutions. That happens because the 
 The right side of Fig. \ref{fig:bernoulli} outlines the loss function during the GD iterations where each colored loss trace corresponds to the associated colored line in the left plots.

% ---------------------------------
% ---------------------------------
% ---------------------------------
% \subsection{Systems of ordinary differential equations: Hamiltonian systems}
{\bf Systems of ordinary differential equations, Hamiltonian systems: }
In this section, we employ   RC to solve systems of ODEs. 
%Higher order ODEs can be decomposed into  systems of first order ODEs, hence solving systems  means that  RC can solve higher order ODEs. %This is a standard procedure followed by standard integrators
To apply  RC to systems, the network architecture needs to be modified to return multiple outputs $N_j$, where $j$ indicates a different output. The number of the $N_j$  needs to be the same with the number of the equations in the system. Each $N_j$ has a different set of weights $\wout^{(j)}$, while all  $N_j$ share the same hidden states, namely:
\begin{align}
  \label{eq:Nsys}
  N_j = \wout^{(j)} \cdot \HH.
\end{align}
Moreover, the loss function (\ref{eq:loss1})  includes all the ODEs included in the system. We exploit the RC solver in solving the equations of motion for a nonlinear Hamiltonian system, the nonlinear oscillator. In this system  the energy is conserved and thus, we adopt the energy regularization introduced in Ref. \cite{mariosHNN} that drastically accelerates the training and improves the fidelity of the predicted solutions. 

% {\bf Hamiltonian systems}
 Hamiltonian systems obey the energy conservation law. Specifically, these systems are characterized by a time-invariant hamiltonian function that represents the total energy. The hamiltonian of a nonlinear oscillator with unity mass and frequency is given by:
\begin{align}
  \label{eq:NL_ham}
  \mathcal{H}(x,p) = \frac{p^2}{2} + \frac{x^2}{2} + \frac{x^4}{4},
\end{align}
and the associated equations of motion read:
\begin{align}
  \label{eq:NL_x} 
  \dot x &= p \\
  \label{eq:NL_p}
  \dot p &= -x - x^3
\end{align}
where $x,~p$ are the position and momentum  variables  \cite{mariosHNN}. The loss function consists of three parts: $L_\text{ODE}$ for the ODEs (\ref{eq:NL_x}), (\ref{eq:NL_p}); a hamiltonian penalty $L_{\mathcal{H}}$ that penalizes violations in the energy conservation and is defined by Eq. (\ref{eq:NL_ham}); and  a  regularization term. Subsequently, the total $L$ is:
\begin{align}
  L &= L_\text{ODE}+ L_{\mathcal{H}} + L_\text{reg} \nonumber \\
   \label{eq:NL_loss}
   &= \sum_{n=0}^{K}
   \Big[ \left(\dot x_n-p_n\right)^2 + \left(\dot p_n + x_n + x_n^3 \right)^2 +\left(E - \mathcal{H}(x_n, p_n)\right)^2 \Big] + L_\text{reg}. %+ \lambda \sum_{j=x,p} \wout^{(j)T} \wout^{(j)}.
\end{align}
$E=\mathcal{H}(x_0, p_0)$ represents the total energy defined by the  ICs $x(0)=x_0$, $p(0)=p_0$. Earlier we choose $L_2$ regularization because we derived  exact solutions for the $\wout$; this was  possible with $L_2$. For systems of ODEs we  do not derive exact $\wout$ and thus, we can apply any $L_\text{reg}$. We use the elastic net regularization which has been shown to be a dominant generalization of $L_1$ and $L_2$ (see the SM) \cite{enet2005}. The RC solutions are defined through Eqs. (\ref{eq:parSol}) and (\ref{eq:Nsys}) as:
\begin{align}
  \label{eq:xn}
  x_n = x_0 + \wout^{(x)} \cdot \left( g(t_n) \HH(t_n) \right),\\
%  \dot x_n = \wout^{(x)} \cdot \left( g(t_n) \HH(t_n) \right)\dot{} \\
  \label{eq:pn}
  p_n = p_0 + \wout^{(p)} \cdot \left( g(t_n) \HH(t_n) \right).
\end{align}

\begin{figure}
  \centering
\includegraphics[scale=0.4]{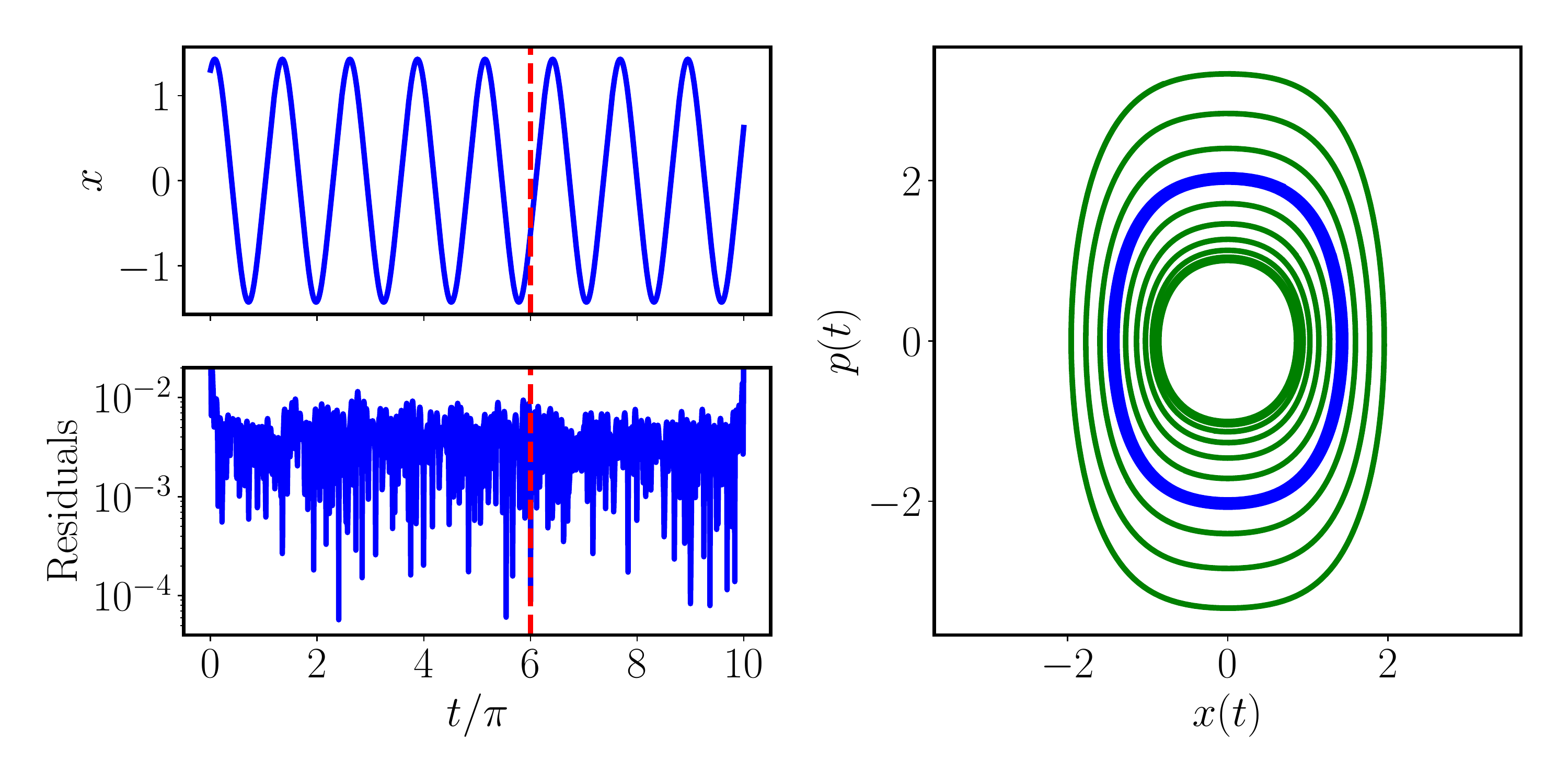}
  \caption{Nonlinear oscillator. Left panel represents a single IC. BO is applied up to red dahsed line. Right graph demonstrate the RC solutions for different ICs not used in BO optimization (green), while blue line is associated to the solution shown in the left plot.  }
  \label{fig:nlOsc}
\end{figure}

We employ the RC solver from the \rctorch library to solve the Eqs. (\ref{eq:NL_x}) and (\ref{eq:NL_p}). Specifically, we minimize Eq. (\ref{eq:NL_loss}) by applying GD and use $g(t)$ of Eq. (\ref{eq:parFun_exp}). The results are outlined in Fig. \ref{fig:nlOsc}. First, we consider a singe set of ICs, $(x_0,p_0)=(1.3, 1)$, and use the hybrid mode consisting of GD and BO to  find the optimal hyper-parameters. This optimization is performed in the time range $t=[0, 6\pi]$. Then, using the obtained hyper-parameters we expand the time range to $t=[0, 10\pi]$ and generate RC solutions  solely using GD. The RC solution for  $x$ is presented in the upper left graph, while the lower plot shows the residuals. In both images, the dashed red line indicates the end of the BO.
Using the same hyper-parameter set, we apply the RC solver in a range of ICs. The right panel we shows the phase-space diagram ($x-p$ plot) for the IC used in the BO (blue line) and for the ICs where only GD is applied (green lines). In the SM we report the residuals and the loss traces during the GD for all the investigated ICs. 
Figure \ref{fig:nlOsc} is evidence that a single hyper-parameter set can be used to solve an ODE system for different ICs and time ranges. 
% Hence, the computationally expensive BO needs is applied only one time. GD is efficient and converges fast as is reported in the SM. 

% \section{Discussion}
\section{Conclusion}
\label{sec:conclusion}
Recently, NN  differential equation solvers have attracted a lot of interest. % from the scientific machine learning community. 
These solvers present some crucial advantages over traditional integrators such as  they  provide analytical and differentiable solutions that can be inverted, suffer less from the "curse of dimensionality", and learn general solutions. %, and the solutions are efficiently stored.
While many methods and scientific libraries for NN solvers have been reported an RNN solver is still missing from the literature.

{\bf Novelty:}
We presented a novel RNN ODE solver in the context of an unsupervised RC and assessed the performance of the RC solver by solving linear and nonlinear ODEs. 
% capable of solving general systems of ODEs. 
We showed that a closed-form solution for the RC weights is possible for solving linear ODEs with explicit time dependence, leading to a backpropagation-free optimization method. For nonlinear system, we applied GD which is very efficient since in the RC architecture we train only a linear output layer. For the hyper-parameter optimization, we employed BO integrated with GD. We found that a single set of hyper-parameters can be shared for solving ODEs for different initial conditions and time intervals. 

{\bf Limitations:}
For  BO we used TURBO-1, however, dominant methods such as TURBO-m have been shown to be more robust and get stuck less often in local minima. 
The efficacy of the proposed RC solver has not been corroborated for very demanding ODEs due to the limited capacity in finding optimal hyperparameters. %Nevertheless, improving the BO method is straight forward. 
Although we derived a closed-form solution for a single linear ODE, such a closed formula has not been derived for linear systems of ODEs. Hence, GD is required for solving systems. 
The proposed RC architecture takes an  input $t$ restricting the RC to solve only ODEs. With more variables as inputs, the RC will be able to solve partial differential equations.% as well as it will be able to learn general solutions, namely solutions as a function of the initial and boundary conditions. 
Currently a uniformly spaced input $t$ has been used, so the efficacy of RC in solving stiff ODEs is not yet determined. 
A thorough investigation of the optimal reservoir dynamics has not been performed in this study. %We use a standard RC architecture. %  and  explored two different activation functions. %
In this work, we demonstrated that unsupervised RC can solve ODEs by solving a few specific problems using the \rctorch library. 
% However, a version of the library which attempts to solve any system of ODEs is under development.
% 
Future updates of \rctorch will remove the weakness discussed above, significantly expanding the potential applications of the library. A more general and more powerful \rctorch version is slated to be released in the coming months.

 Training an RC  is  extremely fast  (see  SM for runtime details and  coding demonstrations). Although more investigation is warranted, RC has the potential to make NN  solvers dramatically faster and even potentially competitive with integrators.
 Furthermore, RC has widely been adopted to form neuromorphic devices. Since a single set of hyper-parameters can be used to solve a system of ODEs for different ranges of ICs and time ranges, a physical reservoir can be designed to respect these hyper-parameters. Then a readout output layer will be efficiently trained to solve a system of ODEs for different ICs and times. Subsequently, the proposed machine learning method can be potentially implemented to form a neuromorphic computing device for solving ODEs.

% ######################################################
% ######################################################
% ######################################################
% ######################################################

\section{Broader Impact}

Solving differential equations is substantial in every scientific field including engineering, applied physics, quantum chemistry, finance, and biology. Solving these equations can be extremely demanding and frequently prohibitive due to the limitations of existing numerical methods. Subsequently, new technologies and more efficient methods for solving differential equations are crucial to accelerate progress in scientific research.
In this work, we introduced a general framework for solving differential equations with recurrent neural networks. We demonstrate the method by solving systems of ordinary differential equations. Yet, this method can be expanded to systems of partial differential equations as well as eigenvalue problems. Moreover, we suggest a computationally efficient method to calculate time derivatives of the outputs of recurrent networks, making possible the development of recurrent data-driven physics-informed neural networks. 

{\bf Societal and Environmental Impact: }
 Solving differential equations with RC may have negative social impacts depending on what the user employs them for, but they are not immediately obvious and are likely indirect.
As far as the environment is concerned,
Bayesian optimization can be computationally expensive, but it is not expensive enough to warrant concerns about environmental impacts when compared to heavier models like transformers, feed forward networks or RNNs such as LSTMs. 
There is an upfront cost with RC, but even after taking this into consideration, our models are likely faster, more efficient, and much less energy intensive (having a smaller carbon footprint) than comparable feed forward neural network differential equation solvers. Moreover, the proposed method can potentially be used to build neuromorphic devices, drastically accelerating computations with extremely low energy consumption. 

%our model is trained much faster that feed forward neural network solvers. Consequently, it demands significantly less energy. 

\section*{Acknowledgments}
 The authors would like to thank  Shaan Desai, Shivam Raval, Hargun Singh Oberoi for their  comments on the manuscript and  numerical experiments. In addition, we would like to thank Reinier Maat for advising us on the development of the \rctorch library.

% \bibliographystyle{unsrt}
% \bibliography{refs}

\printbibliography

\newpage
\section*{Checklist}

\begin{enumerate}

\item For all authors...
\begin{enumerate}
  \item Do the main claims made in the abstract and introduction accurately reflect the paper's contributions and scope?
    \answerYes{}
  \item Did you describe the limitations of your work?
    \answerYes{}. See section: Conclusion.
  \item Did you discuss any potential negative societal impacts of your work?
    \answerYes{}. See section: Broader Impact.
  \item Have you read the ethics review guidelines and ensured that your paper conforms to them?
    \answerYes{}.
\end{enumerate}

\item If you are including theoretical results...
\begin{enumerate}
  \item Did you state the full set of assumptions of all theoretical results?
    \answerYes{}. Sections 3 and 4  have  complete discussions.
	\item Did you include complete proofs of all theoretical results? \answerYes{}. Section 3 and 4  contain all the proofs.
\end{enumerate}

\item If you ran experiments...
\begin{enumerate}
  \item Did you include the code, data, and instructions needed to reproduce the main experimental results (either in the supplemental material or as a URL)?
    \answerYes{}. Details can be found in the supplementary material (SM). We also provide the URL for a github public repository that contains  the   library  and  notebooks used in this study.
  \item Did you specify all the training details (e.g., data splits, hyperparameters, how they were chosen)?
    \answerYes{}. They are stated in section 4 and more details can be found in the SM. 
	\item Did you report error bars (e.g., with respect to the random seed after running experiments multiple times)?
    \answerYes{}. In section 4 and in the SM.
	\item Did you include the total amount of compute and the type of resources used (e.g., type of GPUs, internal cluster, or cloud provider)?
    \answerYes{}. All the details are reported in the SM.
\end{enumerate}

\item If you are using existing assets (e.g., code, data, models) or curating/releasing new assets...
\begin{enumerate}
  \item If your work uses existing assets, did you cite the creators?
    \answerYes{}. 
  \item Did you mention the license of the assets?
    \answerYes{}. 
  \item Did you include any new assets either in the supplemental material or as a URL?
    \answerYes{}. Yes, a new asset for the library used in this study is included.
  \item Did you discuss whether and how consent was obtained from people whose data you're using/curating?
    \answerNA{}. There was no need for consent.
  \item Did you discuss whether the data you are using/curating contains personally identifiable information or offensive content?
    \answerNA{}. In this neural network method we do not use data for the training.  % It is not personally identifiable.
\end{enumerate}

\item If you used crowdsourcing or conducted research with human subjects...
\begin{enumerate}
  \item Did you include the full text of instructions given to participants and screenshots, if applicable?
    \answerNA{}. We did not use crowdsourcing or conducted research with human subjects. 
  \item Did you describe any potential participant risks, with links to Institutional Review Board (IRB) approvals, if applicable?
    \answerNA{}.  We did not use crowdsourcing or conducted research with human subjects. 
  \item Did you include the estimated hourly wage paid to participants and the total amount spent on participant compensation?
    \answerNA{}.  We did not use crowdsourcing or conducted research with human subjects. 
\end{enumerate}

\end{enumerate}

\clearpage

\end{document}

\maketitle

\section*{Supplementary python notebook overview}

We provide 6 python notebooks showing our results presented in the main manuscript and in this supplementary material (SM). They break into two categories: Bayesian optimization (BO) for hyper-parameters used in  the reservoir computing (RC) and for RC training.
The training notebooks demonstrate the differential equation solutions which appeared in the main manuscript and in the SM. The BO notebooks show how we optimized for hyper-parameters for the various models. We note that  BO gives different results after every optimization due to random initialization. Although they can vary in quality, the obtained hyper-parameters are always valid and thus, the BO is robust. All the notebooks also have timed cells so as to approximately reproduce the table \ref{table:1} which shows the computational time taken for the various experiments.

Below is a list of  supplementary jupyter notebooks and a brief description of each. All the notebooks reproducing the results can be found at the following link: \newline \url{https://github.com/blindedjoy/RcTorch/tree/master/final_notebooks}

\begin{itemize}
\item \textbf{Linear\_solutions.ipynb} \newline
Demonstrates the solutions to the linear differential equations (which did not need gradient descent).
\item \textbf{Bernoulli\_solutions.ipynb} \newline
Demonstrates the solutions to the bernoulli equation with three solutions types: linearization only, backpropagation only, and a combination of linearization and backpropagation.
\item \textbf{Systems\_solutions.ipynb} \newline
Demonstrates the solutions to the nonlinear oscillator.
\item \textbf{Linear\_BO.ipynb}\newline
A notebook showing BO similar to what was used to find the hyper-parameter set for the linear equations. Does not need GPUs.

\item \textbf{Bernoulli\_BO.ipynb} \newline
A notebook showing BO similar to what was used to find the hyper-parameter set for the bernoulli equation. Should be run with GPUs.
    
\item \textbf{systems\_BO.ipynb} \newline
A notebook showing BO similar to what was used to find the hyper-parameter set for the nonlinear oscillator equation. Should be run with GPUs.
\end{itemize}

\subsection*{Experiment Runtimes}

In Table \ref{table:1} we report the  experiment times needed for different  experiments presented in the main manuscript and in the SM. All the notebooks are accessible through the link in the above section, hence the experiments are  reproducible. The CPU experiments were run on a MacBook Pro (16-inch, 2019) with a 2.4 GHz 8-Core Intel Core i9 processor and 64 GB 2667 MHz DDR4 of memory. For users with a less powerful machine, a smaller batch size should be used since  the batch size corresponds to the number of RCs being trained in parallel via the multi-processing package; however, this will slow down the computations.
%If you run the experiments on a smaller machine try to use a smaller batch size as the batch size corresponds to RCs being trained in parallel via the multi-processing package, though this will slow down the computations.
On a mac machine, activity monitor should be used to make sure that we  are not overloading the memory which will cause data to be read in and out from the hard-drive which dramatically slows down performance.

The GPU experiments were  run on google colab (with 12 GB of RAM), therefore a batch size of 1 was used due to this constraint. We note that GD (backpropagation) for different initial conditions (ICs) could be run in parallel, however,  this is not yet implemented in the current version of the \rctorch. In addition, the library has not been fully optimized implying large further optimization gains.

\begin{table}[h!]
\centering
\begin{tabular}{c c c c c c c} 
    \toprule
    {Differential equations} & {\# nodes} & {\# ICs} & {solve time} & {BO time} &  {epochs} & {batch size} \\ \midrule
    {\textbf{Linear ODEs (CPUs)}} &  & {20} &  & & \\
    simple population  & 250 & 20 & 0.795 & 184 & {--} & 10 \\
    driven population & 500  & 20 & 1.62  & 254 & {--} & 10\\
    $t-$dependent coefficients  & 500  & 20 & 0.895 & 197 &  {--} &10 \\\midrule
    \textbf{Linear ODEs (CPUs)} &  & {100} &  & & & \\
    simple population  & 250 & 100 & {1.32} & {316} & {--} & 10 \\
    driven population & 500  & 100 & 3.74 & 864 & {--} & 10 \\
    $t-$dependent coefficients & 500  & 100 & 3.03  & 774 &  {--} &10\\\midrule
    \textbf{Nonlinear Bernoulli (GPU)} &  &  &  & & \\
    Linearization  & 500  & 4 & 1.15 & --  & -- & --   \\ 
    GD only  & 500  & 4 & 191 & 3075 & 5000 & 1    \\ 
    GD with linearized weights  & 500 &4 & 184 &  3075 & 5000  & 1   \\ 
    \\\midrule
    \textbf{System of ODEs (GPU)} &  &  &  & & \\
    Nonlinear oscillator  & 500   & 2 & 129  & 4440 & 5000 & 1  \\ \bottomrule\\
\end{tabular}
\caption{Computational time needed by the RC solver for the experiments presented in the notebooks   }
\label{table:1}
\end{table}

% \pagebreak
\subsection*{Hyper-parameters list}
 For all experiments a random state of 209 was used which was set with pytorch's \texttt{manual\_seed} code. We did not optimize the random state.
% \newline
% \begin{minipage}{\linewidth}
\begin{lstlisting}
Simple population equation (not presented in the main manuscript):
{'dt': 0.0031622776601683794,
 'n_nodes': 250,
 'connectivity': 0.7170604557008349,
 'spectral_radius': 1.5755887031555176,
 'regularization': 0.00034441529823729916,
 'leaking_rate': 0.9272222518920898,
 'bias': 0.1780446171760559}

Driven population equation (Eq. (23) in the main manuscript):
{'dt': 0.0031622776601683794,
 'n_nodes': 500,
 'connectivity': 0.7875262340500385,
 'spectral_radius': 9.97140121459961,
 'regularization': 8.656278081920211,
 'leaking_rate': 0.007868987508118153,
 'bias': -0.2435922622680664}

First order ODE with time dependent coefficients (Eq. (25) in the main manuscript):
{'n_nodes': 500,
 'connectivity': 0.09905712745750006,
 'spectral_radius': 1.8904799222946167,
 'regularization': 714.156090350679,
 'leaking_rate': 0.031645022332668304,
 'bias': -0.24167031049728394,
 'dt' : 0.005}

Bernoulli type nonlinear ODE (Eq. (26)  in the main manuscript): 
{'dt': 0.007943282347242814,
 'n_nodes': 500,
 'connectivity': 0.0003179179463749722, 
 'spectral_radius': 7.975825786590576, 
 'regularization': 0.3332787303378571,
 'leaking_rate': 0.07119506597518921,
 'bias': -0.9424528479576111}
epochs: 30000
learning_rate: 0.01
spikethreshold: 0.25


Nonlinear Oscillator system (Eqs. (34) and (35)  in the main manuscript):
{'dt': 0.001,
 'regularization': 48.97788193684461,
 'n_nodes': 500,
 'connectivity': 0.017714821964432213,
 'spectral_radius': 2.3660330772399902,
 'leaking_rate': 0.0024312976747751236,
 'bias': 0.37677669525146484,
 'enet_alpha': 0.2082211971282959,
 'enet_strength': 0.118459548397668,
 'spikethreshold': 0.43705281615257263,
 'gamma': 0.09469877928495407,
 'gamma_cyclic': 0.999860422666841}
epochs: 50000
learning_rate: 0.01
\end{lstlisting}
% \end{minipage}

\section*{Numerical implementation}

% {\bf RC architecture:}
\subsection*{RC architecture}
The \rctorch library  presented and used in the study was written in pytorch and was adapted from the library developed in Ref. \cite{Maat2018}. 
We used RC reservoirs consisting of 500 hidden nodes. This architecture was found to be sufficiently expressive to solve the ODEs discussed in the main manuscript of this study. We note that  the computational complexity and memory cost of an RC grow quadratically with respect to the number of nodes \cite{dong2020reservoir}. %, which justifies what is a relatively small reservoir. 
In $\rctorch$ we are allowed to choose any activation function we want as long as it is bounded; unbounded activation functions such as \emph{ReLU} have been shown not to converge in the context of RC \cite{dong2020reservoir}. For the first order ODEs we use $\tanh(\cdot)$ activation, whereas, for the system of ODEs we apply a $\sin(\cdot)$ activation since it has been shown that in dynamical systems this type of activation is very effective \cite{mariosHNN}. 
%Sin is an excellent choice of activation for higher derivatives due to the tame nature of the derivatives and in addition only bounded activation functions will converge with RC \cite{dong2020reservoir}.

\subsection*{Bayesian optimization}
% {\bf Bayesian optimization, Turbo-1 and Thompson Sampling overview:}
Although RNNs are very powerful models in the handling of sequential data, they suffer from {the vanishing and exploding gradient problem}. Echo-state RNNs like RCs eliminate this issue by fixing their input, hidden weights, and hidden and biases.  Moreover, training RNNs over long sequences (time series) is a computationally demanding task. Echo state networks, by contrast, are very efficient and extremely fast due to their special architecture. The computationally expensive aspect of building an effective RC model is to find the optimal hyper-parameters that define the network architecture. Here, we review a BO method that is introduced in Ref. \cite{Maat2018}  and is adopted in the library (\rctorch)  used in this study.
%Since the training of the RC consists of training just one linear layer, the training is extremely fast. However, the RC is extremely sensitive to the hyperparameters. Hence, finding the optimal set of hyperparameters is a crucial part for having a well-trained RC network. Grid search has been widely  used to tune a few hyperparameters while the other hyperparameters are fixed. A more efficient optimization was introduced in \cite{Maat2018} that employes Bayesian optimization (BO). 
%This  approach makes it possible to search for many hyper-parameters in a high-dimensional space. 
%Similar approaches based on BO have been recently used and comprise an active field of research CITATIONS. In this study we employ the method introduced in \cite{Maat2018}. 
%In particular, we have developed a python library called \textsc{rcTorch} CITATION AND GIT that provides an RC network along with BO for fast optimal hyper-parameter tuning. 
%Review the reservoir computing formulation.  Say about the sensitivity to the hyper-parameters. Describe the Bayesian optimization. 

{\bf Turbo-1: }
BO is a powerful method for  analyzing expensive acquisition functions like the training of neural networks. It is a likelihood maximization problem where the surrogate model is typically a gaussian process model (GP). We use the Turbo-1 algorithm outlined in a BoTorch tutorial \cite{TURBO2020, Botorch2020} which,  like global BO methods, is robust to noisy observations and has rigorous uncertainty estimates. However, an advantage of Turbo over other BO algorithms is that it does not  suffer from "the overemphasized exploration that results from global optimization" and does not "scale poorly to high dimensions", problems common observed problem with GP models \cite{TURBO2020}. That is, GP tends to be good at exploring the hyper-parameter space but it does not perform well at optimizing in local regions. Turbo performs well by breaking down the global optimization problem into many smaller local optimization problems.  Moreover, Turbo has been shown to be faster than other state-of-the-art black box evaluation methods (BO as well as other types of methods) and has been shown to find better global maxima. Currently, \rctorch library embeds Turbo-1, however, an upgrade to Turbo-m is scheduled and will make the hyper-parameter search more robust and less likely to get stuck in local maxima than Turbo-1. 
% However,  Turbo-1 is  is fairly random and can still get stuck in local maxima. For example, when preparing this SM post submission we came across a much stronger Bernoulli result than the one in the main paper. You can find these hyper-parameters in the supplemental Bernoulli$\_$BO.ipynb notebook.

{\bf Thompson Sampling: }
Furthermore, we use a Thompson sampling (TS) acquisition function since it has been shown (unlike many other common acquisition functions) to scale linearly in terms of complexity with increased batch size \cite{TURBO2020}. \rctorch trains multiple RCs in parallel for each batch and larger batch sizes have shown to have better performance in combination with Turbo and TS. In particular, Turbo has been shown to work well with large batch sizes. Larger batch sizes can result in better solutions while not causing a dramatic increase overall TURBO runtime (which is a huge problem for other BO methods). In addition, TS is an efficient acquisition function for exploration which complements Turbo's relative exploitation strength. Thus, the BO used in \rctorch is a modern and balanced technique. The combination of TURBO and TS are excellent algorithmic choices which make \rctorch very efficient  and hold promise for the application of \rctorch on more complex, higher dimensional, and potentially chaotic problems in the future. In addition, these algorithms are well suited for large numbers of hyper-parameters and more robust solutions making it likely that \rctorch will be able to efficiently solve problems with dozens or even hundreds of hyper-parameters in the near future.

{\bf BO cross validation:}
During every iteration of BO with \texttt{rcTorch} we evaluate our model based on a specified number of cross validation samples $n$. This number plays a role in our BO loss function in Eq. (\ref{eq:BOloss}). Each cross validation sample corresponds to the training of one RC network with one set of hyper-parameters.
First BO randomly selects the start of a subset of the training range with the \texttt{subsequence\_length} argument which determines how many time points make up the cv\_sample. Every \texttt{cv\_sample} has a training set and a validation set immediately following the training set in sequence. This step of BO is performed to make BO more robust which has been demonstrated in \cite{Maat2018}.  Next the sub-sequence is split into training and validation sets with proportions determined by the \texttt{val\_split} argument. If the \texttt{val\_split}$~\simeq 0.3$ then 70\% of the sub-sequence length would be the training set and 30\% would be the validation set. The weights of the output layer are optimized based on the training set and then we evaluate the RC on both the training set and the validation set. The validation set corresponds to an evaluation of the models ability to extrapolate, though this requires further study and improvement. %We  initialized the reservoir hidden weights using a normal random distribution.  %Nevertheless further exploration into reservoir dynamics  for possible improvements in extrapolation.
%used a  random uniform reservoir (and did not optimize the random state) so our lab is beginning to conduct an exploration into reservoir dynamics for possible improvements in extrapolation.

{\bf BO loss function:}
In this study we introduce an unsupervised RC where the only input is evenly spaced time points. 
%Because this input does not include data (it is a data-less process), we cannot cheat by including the training set in our evaluation of model performance when solving ODEs with \rctorch. 
%
 In general performing BO for bundles of ICs was more robust than simply using one IC. The second sum of the loss function in Eq. (\ref{eq:BOloss})  corresponds to the ICs considered to solve an ODE. It is worth noting that solving an ODE for different ICs requires one set of the hidden states, since one set of hyper-parameters is sufficient. This property reduces the computation cost since we do not need to construct the hidden states for every different IC. This allows \rctorch to optimize  for hyper-parameters which are robust over a range of ICs. These can be randomly sampled at every BO iteration or alternatively can a consistent list of ICs can be evaluated at every step. We selected the determinstic IC bundle method though the sampling IC method deserves further study. We can use up to $m$ initial conditions yielding the loss function: 

\begin{align}
    \label{eq:BOloss}
\mathcal{L}(t, N, \dot{N}) = \frac{1}{n ~ m} \sum_{ \text{cv} = 1}^n \sum_{\text{IC} = 1}^m \left( \beta \log(\mathcal{L}_\text{train}(t, N, \dot{N})) + (1-\beta)\log(\mathcal{L}_\text{val}(t, N, \dot{N}))\right),
\end{align}
where the BO level hyper-parameter is set to be $\beta=0.5$, $N$ is the RC output, and  $\dot N$ the time derivative of the RC output.

Because we are using an unsupervised RC without having labeled data, we can evaluate it on both the training and validation sets. By contrast, when using \rctorch for regression (where we have labeled-data) we cannot evaluate the RC on the training set as this would lead to over-fitting. 

{\bf Bayesian optimization hyper-parameter descriptions:}
Here, we describe the key  hyper-parameters used in the BO of \rctorch.

 \texttt{Spectral Radius}: The largest eigenvalue of the adjacency matrix governing the reservoir connections \cite{Ott2017}.

\texttt{Connectivity}: The proportion of reservoir connections (weights) that are non-zero. This hyper-parameter determines  the sparsity of the reservoir, which is  a key property which leads to the echo-state property \cite{Jaeger2004}.

\texttt{dt}: The discrete time step determining the step between two sequential  input data-points. This directly determines the behaviour of the reservoir states. Equally-spaced input points are considered through this study. %yielding a universal.

\texttt{leaking\_rate}: This hyper-parameter determines the memory of the reservoir and analogous to similar parameters observed in other leaky recurrent units such as GRUs.

\texttt{regularization}: The ridge  regularization strength  used when solving the linear equation for the exact solutions of the RC trainable parameters. % Also denoted by  $\lambda$ in the literature.

\texttt{n\_nodes}: The number of reservoir nodes (denoted by $M$ in the main manuscript). This hyper-parameter determines the dimensions of the adjacency matrix and the hidden states. More nodes means more expressivity and also more memory usage yielding a slower model.

\texttt{bias}: The bias term of the hidden states update.

\subsection*{RC Training}

{\bf Gradient descent: }
The only layer that is trained in the RC is a linear output layer characterized by $\wout$. Considering a loss function that consists of a mean-square-error (MSE) part and a ridge regularization, we have to minimize a convex  loss function $L$. In the linear ODEs we showed that an exact solution for the $\wout$ that minimizes the loss function is possible. This is not possible for nonlinear problems. However, we can use gradient descent (GD) to minimize the loss function. Since we are dealing with a convex loss function, the GD  is expected  to be very fast. This study provides evidence of such speed.
 For nonlinear ODEs where we lack an exact solution we employ \texttt{StepLR}, a pytorch learning rate scheduler to improve GD. We start with high learning rates in the range $[0.04,  0.01]$, and allow the RC to monitor the loss functions for "spikes" that exceeded some threshold $\psi$ with typical value around 0.1. If $\psi > L_n - L_{n-1}$, where $n$ is the iteration index in the GD and $L$ denotes the loss function, we allow the optimizer to reduce the learning rate by a factor $\gamma=0.1$. Because of these spikes, we keep the best weights observed over the loss period, not the weights observed at the final iteration. In addition, for the second order ODEs we introduce pytorch's \texttt{CyclicLR} with option \texttt{exp\_range} with a corresponding hyper-parameter $\gamma_\text{cyclic}\in [0.99, 0.9999]$. When employed for 5000 iterations, the cyclic learning rate was found, based on empirical evidence, to dramatically accelerates GD and improves the quality of the final solutions. In particular, it was found to rapidly move to the convex part of the loss and help the GD  to avoid local minima. This was particularly true for the non-linear oscillator system where both $\gamma$ and $\psi$ were introduced as hyper-parameters to the nonlinear ODEs during BO. At every iteration (epoch) the cyclic learning rate scheduler decays the learning rate by $\gamma_\text{cyclic}$. However, if the cyclic learning rate were employed for later GD iterations (>5000) we observed that loss may explode and not recover. That is, the loss might experience dramatic spikes (on the order of $10^6$) thereby destroying GD. Further investigation into the cyclic learning rate is warranted.

{\bf Elastic Net: }
In this study we used elastic net regularization for the non-linear oscillator. In particular, we introduced two hyper-parameters associated with elastic net: \texttt{enet\_alpha} and \texttt{enet\_strength}. Elastic net regularization is a combination of L1 and L2 regularization and it is common practice to use the following formula. Let $\rho$ represent \texttt{enet\_alpha} and $\omega$ represent \texttt{enet\_strength}, the the loss function reads
\begin{align*}
  L = \omega~ (\rho  L_1 + (1-\rho)~L_2 ),  
\end{align*}
$\rho$ can be seen as the L1 regularization proportion of the elastic net term while $(1-\rho)$ can be seen as the L2 regularization proportion. $\omega$ is the overall strength of this term. We searched for $\rho$ between 0 and 1 and for $\omega$ on a logarithmic scale.

%enet_strength*(enet_alpha * weight_size_sq + (1- enet_alpha) * weight_size_L1)

%\section*{Hybrid Advantage}
\section*{Linearized RC output weights}
\begin{figure}[h!]
    \centering
    \includegraphics[scale=0.4]{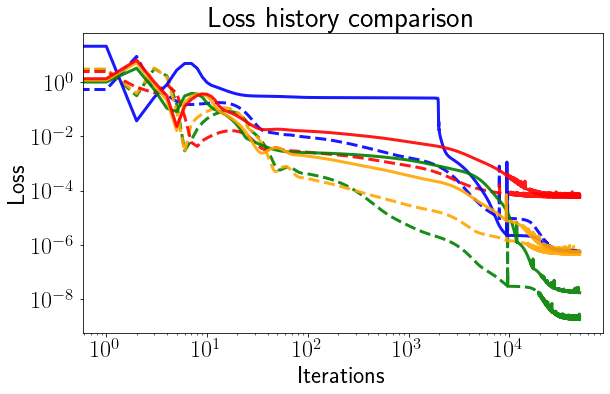}
    \caption{Loss trace during gradient descent for the nonlinear Bernoulli ODE (Eq. (26) in the main manuscript). The solid lines correspond to a randomly initialized output layer weight matrix $\wout$; the dashed lines represent the training loss when linearized $\wout$ are used as initial weights. Each color line corresponds to a different initial condition. } 
    \label{fig:bernoulli}
\end{figure}

We present a numerical exploration to show how the linearization approach discussed in the main manuscript improves the RC training during GD optimization. A linearized closed-form formula for the $\wout$  is derived and given in the main manuscript by Eq. (31). Here, we apply the RC solver to the nonlinear bernoulli ODE (Eq. (26) in the main text). In  Fig. \ref{fig:bernoulli} we report the loss function during GD iteration for the 4 initial conditions  shown in different colors (one for each IC); we explore the same initial conditions  investigated in the main manuscript. In particular, the solid lines in Fig. \ref{fig:bernoulli} show  the loss trace when $\wout$ are randomly initialized whereas, the dashed curves represent the GD training when the $\wout$ are initialized by linearized formula (Eq. (31) in the main text).
We observe in Fig. \ref{fig:bernoulli} that starting with the linearized $\wout$  the loss functions converge faster and reach lower values (overall lower minima in the loss) than starting from random initial weights. To see evidence of the overall minima (as the Fig. \ref{fig:bernoulli}  shows overlapping lines for many trajectories in the final iterations  of training),   see the bernoulli solutions notebook in the github repo.

%\mm{to be removed}
%\includegraphics[scale=0.35]{figures/SM_strongest_bernoulli.png}
\section*{Nonlinear Oscillator}
\begin{figure}[h!]
    \centering
    \includegraphics[scale=0.4]{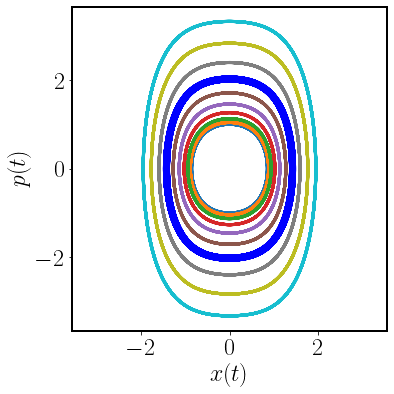}
    \includegraphics[scale=0.4]{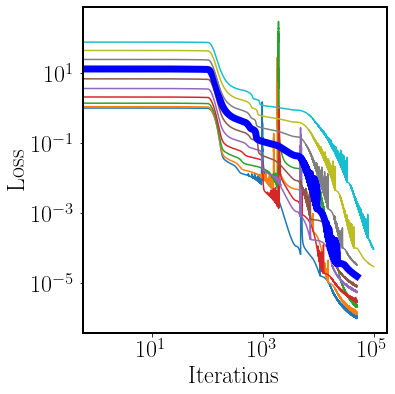}
    \includegraphics[scale=0.4]{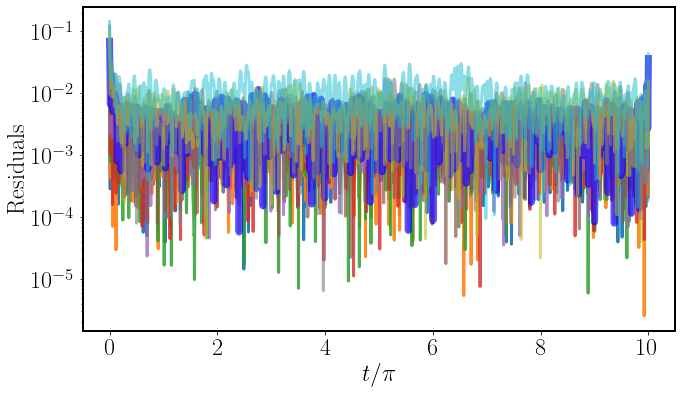}
    \caption{Caption}
    \label{fig:nl_osc}
\end{figure}

In this section, we present more  details about the training and the performance of the RC solver when it is applied to the nonlinear oscillator.  Figure \ref{fig:nl_osc}  presents the results for different different ICs where each IC is represented by a different color; the thicker blue line corresponds to the only IC used in the BO. More specifically, in the left upper diagram we show the  RC prediction for the phase space, namely  an $x-p$ plot. The right graph represents the loss function during the GD iterations. The lower panel outlines the residuals for the RC solution solutions. % We observe that  the GD becomes more demanding for larger  ICs. We intuitionally  expect this behavior since the larger the ICs the more the nonlinear behavior becomes dominant.

\section*{Comparison between RC ODE solver and forward Euler integrator }
In this section we compare the numerical solution speed and  accuracy obtained by the RC solver with the solutions obtained by a Euler integrator. For a fair comparison, in both solvers we used the same $dt$ and  computed 40  initial conditions in the range $[-10,10]$; all the experiments were performed on the same machine. We explore and present the results  for two first order linear ODEs. The simple population equation
\begin{align}
    \label{eq:simplePop}
    \dot y + y = 0,
\end{align}
and the driven population equation (Eq. (23) in the main manuscript)
\begin{align}
    \label{eq:drivenPop}
    \dot y + y = \sin(t).
\end{align}

\begin{figure}[h!]
    \centering
    \includegraphics[scale=0.4]{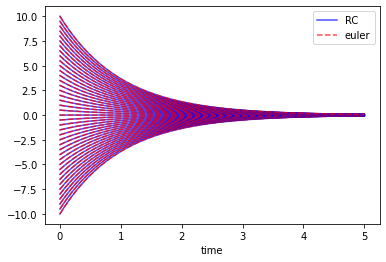}
    \includegraphics[scale=0.4]{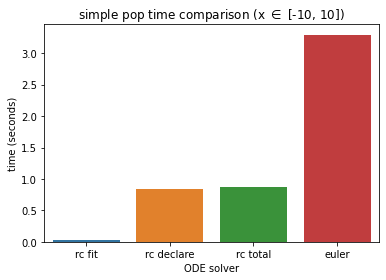}
    \includegraphics[scale=0.4]{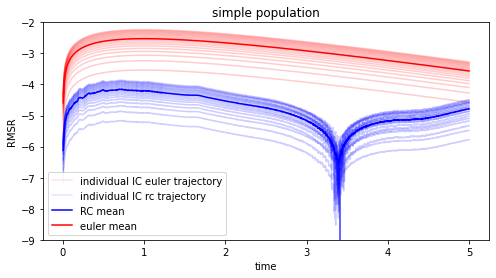}
    \caption{Comparison between RC and Euler ODE solvers one the  population of Eq. (\ref{eq:simplePop}). The top left plot outlines the predicted solutions and the right graph represents the total computational time used for each method. The lower image shows the root mean squared error of the predicted numerical solutions.}
    \label{fig:comparison_simplePop}
\end{figure}

\begin{figure}[h!]
    \centering
    \includegraphics[scale=0.4]{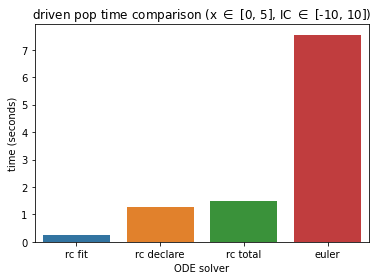}
    \includegraphics[scale=0.4]{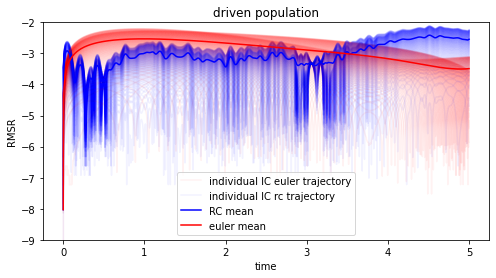}
    \caption{Comparison between RC and Euler ODE solvers one the driven  population of Eq. (\ref{eq:drivenPop}). Left: The total computational time used for each method. Right: The residuals of the predicted numerical solutions.
    }
    \label{fig:comparison_drivenPop}
\end{figure}

For the RC solver, we do not include the time needed for the BO, which is about 3 minutes as shown by Table \ref{table:1}. Moreover, since we solve  linear ODEs, we do not apply GD since there is a closed-form formula for the RC training. In this context, we observe in Figs. \ref{fig:comparison_simplePop} and \ref{fig:comparison_drivenPop} that RC solver is significantly faster than Euler integrator. "RC declare" refers to the time take to declare the esn object, build the hidden states, and solve for the first IC. "RC fit" is the time taken to solve for the remaining ICs.

% % \includegraphics[scale=0.5]{figures/simple_pop_pred.png}
% We will delete this, but it's here for you to see
% % \includegraphics[scale=0.5]{figures/simple_pop_acc.png}
% % \includegraphics[scale=0.5]{figures/simple_pop_time.png}
% % \includegraphics[scale=0.5]{figures/driven_time_comp.png}
% % \includegraphics[scale=0.75]{figures/driven_pop_acc.png}